\documentclass[sigconf,authorversion]{acmart}

\AtBeginDocument{%
  \providecommand\BibTeX{{%
    \normalfont B\kern-0.5em{\scshape i\kern-0.25em b}\kern-0.8em\TeX}}}

\usepackage{multirow}
\usepackage{subcaption}
\usepackage[ruled,vlined]{algorithm2e}
\usepackage{amsthm}
\newtheorem*{theorem*}{Theorem}

\copyrightyear{2021}
\acmYear{2021}
\setcopyright{acmlicensed}
\acmConference[MIG '21]{Motion, Interaction and Games}{November 10--12, 2021}{Virtual Event, Switzerland}
\acmBooktitle{Motion, Interaction and Games (MIG '21), November 10--12, 2021, Virtual Event, Switzerland}
\acmPrice{15.00}
\acmDOI{10.1145/3487983.3488301}
\acmISBN{978-1-4503-9131-3/21/11}

\begin{document}

\title{PFPN: Continuous Control of Physically Simulated Characters using Particle Filtering Policy Network}

\author{Pei Xu}
\email{peix@clemson.edu}
\affiliation{%
  \institution{Clemson University}
  \city{Charleston}
  \state{South Carolina}
  \country{USA}
}

\author{Ioannis Karamouzas}
\email{ioannis@clemson.edu}
\affiliation{%
  \institution{Clemson University}
  \city{Charleston}
  \state{South Carolina}
  \country{USA}
}
\renewcommand{\shortauthors}{Pei Xu and Ioannis Karamouzas}

\begin{abstract}
Data-driven methods for physics-based character control using reinforcement learning have been successfully applied to generate high-quality motions. 
However, existing approaches typically rely on Gaussian distributions to represent the action policy, which can prematurely commit to suboptimal actions when solving high-dimensional continuous control problems for highly-articulated characters.
In this paper, to improve the learning performance of physics-based character controllers, we propose a framework that considers a particle-based action policy as a substitute for Gaussian policies. 
We exploit particle filtering to dynamically explore and discretize the action space, and track the posterior policy represented as a mixture distribution.
The resulting policy can replace the unimodal Gaussian policy which has been the staple for character control problems, without changing the underlying model architecture of the reinforcement learning algorithm  used to perform policy optimization. 
We demonstrate the applicability of our approach on various motion capture imitation tasks. 
Baselines using our particle-based policies achieve better imitation performance and speed of convergence as compared to corresponding implementations using Gaussians, and are more robust to external perturbations during character control.
Related code is available at: {\tt https://motion-lab.github.io/PFPN}.
\end{abstract}



\begin{teaserfigure}
  \includegraphics[width=0.495\linewidth]{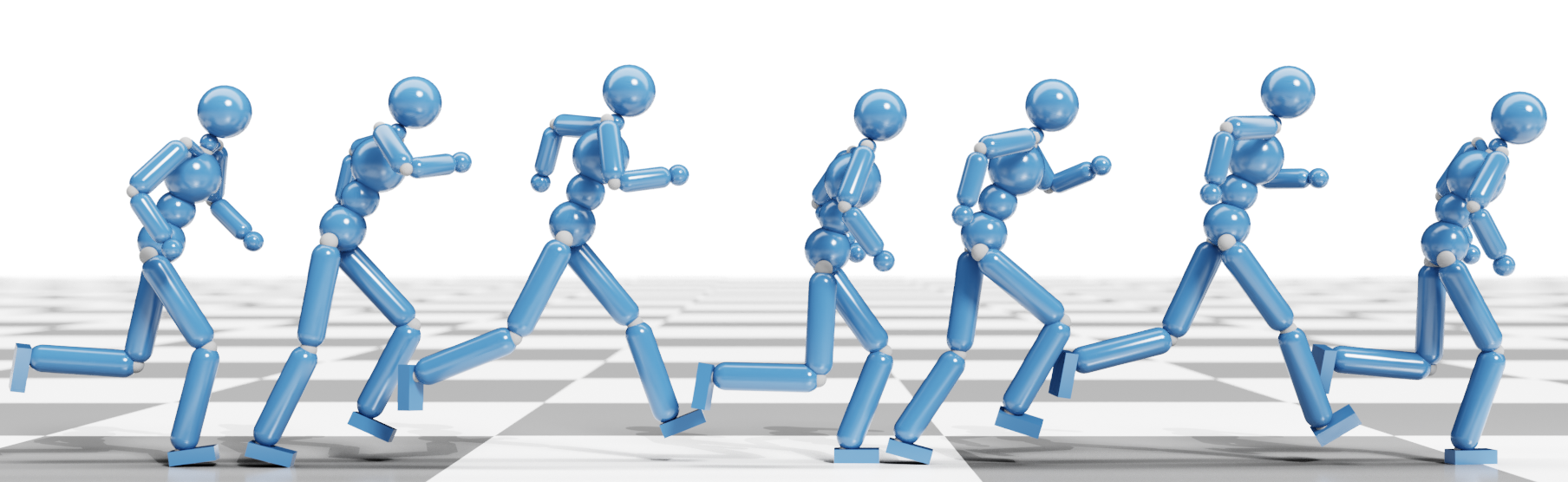}
  \hfill
  \includegraphics[width=0.495\linewidth]{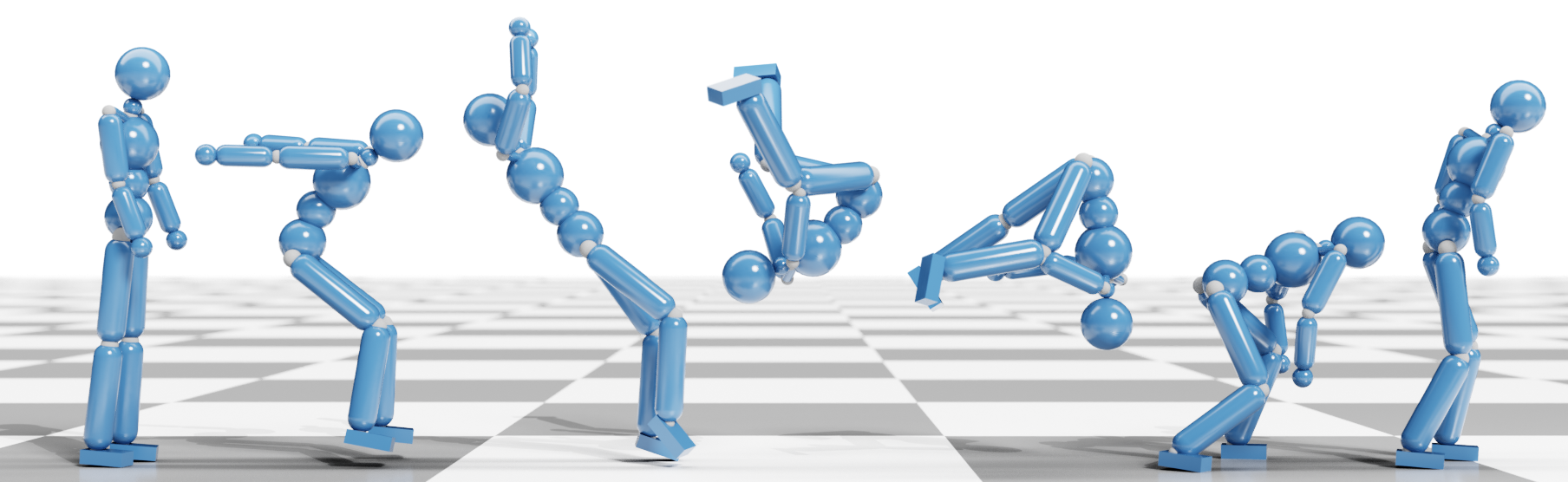}
  \includegraphics[width=0.495\linewidth]{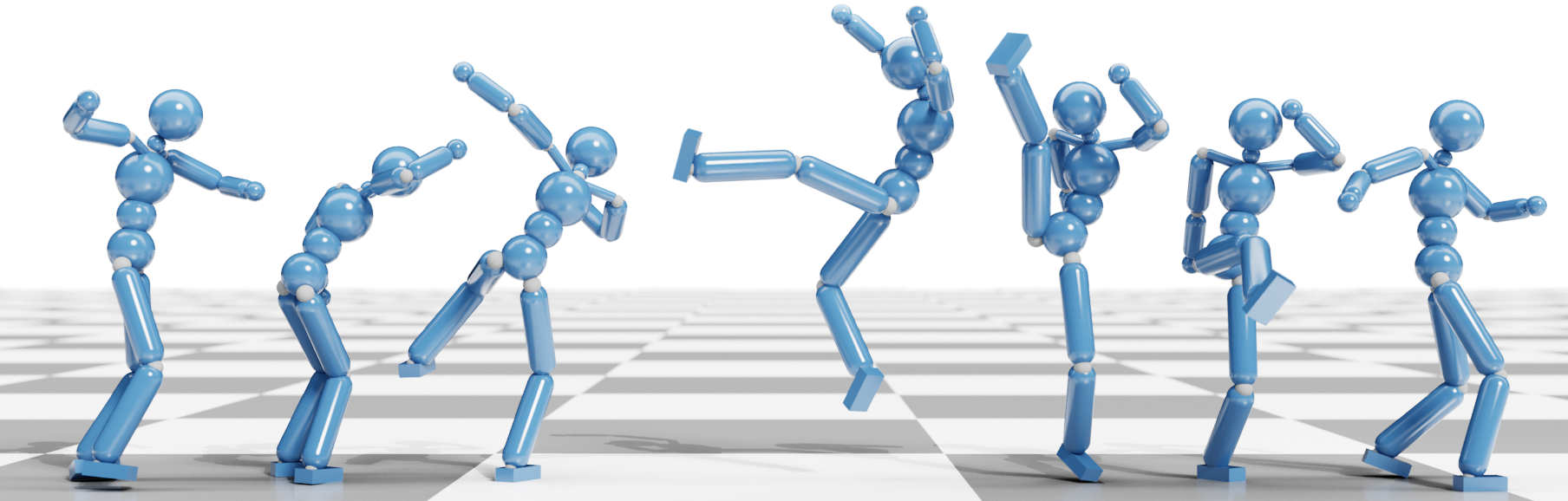}
  \hfill
  \includegraphics[width=0.495\linewidth]{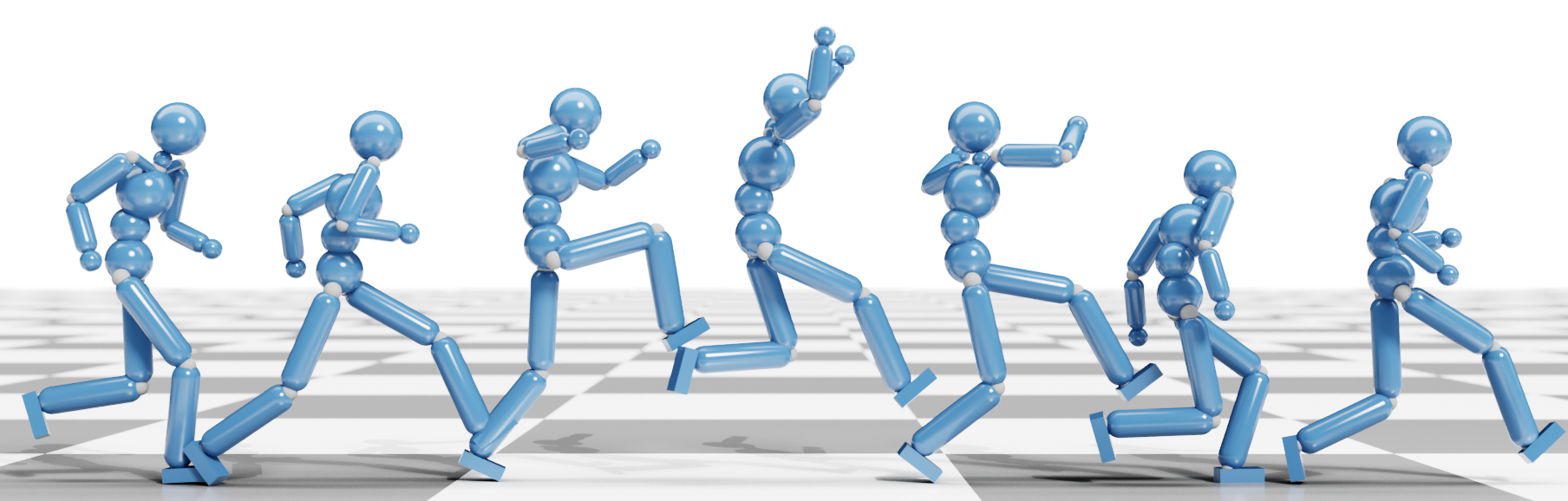}
  \includegraphics[width=\linewidth]{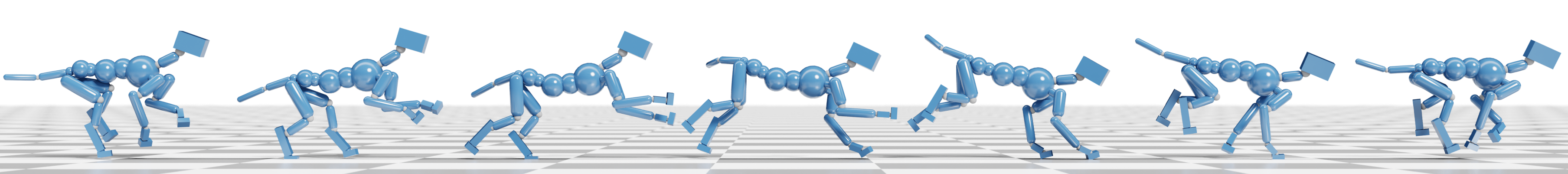}
  \caption{Motions generated through imitation learning using Particle Filtering Policy Network.}
  \label{fig:teaser}
\end{teaserfigure}

\maketitle

\section{Introduction}
In the last few years, 
impressive results have been obtained for physics-based character control using data-driven methods under the framework of deep reinforcement learning (DRL)~\cite{peng2018deepmimic,merel2017learning,nuttapong,ScaDiver,bergamin2019drecon,ma2021learning,yin2021discovering}.
Works from~\citep{peng2017learning,reda2020learning,lee2019scalable,jiang2019synthesis} studied the learning and control performance with different action spaces using joint or muscle actuation formulations, and show the impact that action parameterization has on the trained policies.
State-of-the-art approaches typically perform exploration in the action space of joint angles and control the character using PD servos to generate high-fidelity motions through imitation learning.
Despite recent success, though, generating high-quality and robust animation for highly articulated characters is still a challenging task. Due to the infinite feasible action choices, controlling many degrees of is inherently ambiguous with respect to most behaviors, resulting in control problems that are under specified and highly dimensional.

State-of-the-art approaches for continuous character control define the action policy as a multivariate Gaussian distribution with independent components.
Nevertheless,
the unimodal form of a Gaussian distribution could 
prematurely commit to suboptimal actions when optimizing a reward function that consists of multiple competing terms and has a multimodal landscape~\cite{hamalainen2020visualizing}. 
Consider, for example, an articulated character that needs to learn a mapping from states to actions based on all individual joints while tracking a reference motion. This is a very challenging task as for a given DOF and a continuous action space, we need to find a state-dependent mean and standard deviation. 
This process can be imagined as sliding the Gaussian 
to determine where to place it on an infinite line while also shrinking and swelling the distribution. And of course the problem becomes even more complicated as we need to do the same thing for all DOFs and determine how they work in synchrony for a given state.

To improve the policy expressivity beyond the unimodal Gaussian, 
recent works in character animation use primitive actions~\cite{peng2019mcp}, coactivations~\cite{ranganath2019low}, or a mixture of experts~\cite{ScaDiver} 
though the underlying distribution is still Gaussian.
To address the unimodality issue of Gaussian policies, in the field of DRL, people have been exploring more expressive distributions than Gaussians, 
with a simple solution being to discretize the action space and use categorical distributions as multi-modal action policies~\cite{andrychowicz2020learning,jaskowski2018reinforcement,tang2019discretizing}. 
However, as we show, such a solution 
cannot scale well 
when training expert-guided control policies for physically simulated characters.   
The reason is that the performance of the action-space discretization  
depends a lot on the choice of discrete atomic actions, which are usually picked uniformly due to lack of prior knowledge. Therefore, a simple, fixed action discretization scheme typically can only provide suboptimal solutions that are unable to meet the fine control demands. 

In this paper, we introduce an expressive, multimodal action policy for 
DRL-based learning of physics-based controllers for highly articulated characters. 
Following prior work, we employ a joint-actuation space and learn a mapping from states to joint target angles that are given as input to PD servos for computing the corresponding torque values.
Instead of representing each action as a state-depending Gaussian, though, we exploit a particle approach 
to dynamically sample the action space during training and
track the policy represented as a mixture of Gaussian distributions with state-independent components. 
We refer to the resulting policy network as Particle Filtering Policy Network (PFPN).
We evaluate PFPN on benchmarks from the DeepMimic framework~\cite{peng2018deepmimic,peng2020learning} involving  motor control tasks for a humanoid and a dog character (see Figure~\ref{fig:teaser} for some results).
Our experiments show that baselines using PFPN exhibit better imitation performance and/or speed of convergence as compared to state-of-the-art Gaussian baselines, and lead to more robust character control under external perturbation. 
In addition, PFPN-trained controllers
can generate motions with high visual quality for
articulated characters
performing motor tasks with the grace and naturalness of complex beings.  
Overall, PFPN offers a great alternative to Gaussian action policies leading to state-of-the-art controllers for physically simulated characters
without introducing any notable computational overhead. 

\section{Background}
We consider a standard reinforcement learning setup where given a time horizon $H$ and the trajectory \mbox{$\tau = (\mathbf{s}_1, \mathbf{a}_1, \cdots, \mathbf{s}_H, \mathbf{a}_H)$} obtained by a transition model $\mathcal{M}(\mathbf{s}_{t+1} \vert \mathbf{s}_t, \mathbf{a}_t)$ and a parameterized action policy $\pi_\theta(\mathbf{a}_t \vert \mathbf{s}_t)$, with $\mathbf{s}_t$ and $\mathbf{a}_t$ denoting the state and action taken at time step $t$, respectively, the goal
is to find the policy parameters 
$\theta$ that maximize the cumulative reward:
\begin{equation}
J(\theta) = \mathbb{E}_{\tau \sim p_{\theta}(\tau)} \left[r(\tau)\right] = \int p_{\theta}(\tau) r(\tau) d\tau. 
\end{equation}
Here, $p_{\theta}(\tau)$ denotes the state-action visitation distribution for the trajectory $\tau$ induced by the transient model $\mathcal{M}$ and the action policy $\pi_\theta$, 
and $r(\tau)=\sum_tr(\mathbf{s}_t,\mathbf{a}_t)$ where $r(\mathbf{s}_t,\mathbf{a}_t)$ is the reward received at time step $t$. 
We can maximize $J(\theta)$ by adjusting the policy parameters $\theta$ through the gradient ascent method, where the gradient of the expected reward can be determined according to the policy gradient theorem~\cite{sutton2000policy}, i.e.
\begin{equation}
\label{eq:policy_gradient}
    \nabla_\theta J(\theta) = \mathbb{E}_{\tau \sim \pi_\theta(\cdot|\mathbf{s}_t)} \left[A_t \nabla_\theta \log \pi_{\theta}(\mathbf{a}_t|\mathbf{s}_t) \vert \mathbf{s}_t \right],
\end{equation}
where $A_t$ denotes an estimate to the reward term $r_t(\tau)$.
In DRL, the estimator of $A_t$ often relies on a separate network (critic) that is updated in tandem with the policy network (actor). This gives rise to a family of policy gradient algorithms known as actor-critic. 

Given a multi-dimensional continuous action space, the most common choice in current DRL baselines is to model the policy $\pi_\theta$ as a multivariate Gaussian distribution with independent components for each action dimension.
For simplicity, let us consider a simple case with a single action dimension and define the action policy as $\pi_{\theta}(\cdot|\mathbf{s}_t) := \mathcal{N}(\mu_\theta(\mathbf{s}_t), \sigma_\theta^2(\mathbf{s}_t))$. 
Then, we can obtain
$\log \pi_{\theta}(a_t|\mathbf{s}_t) \propto -(a_t-\mu_\theta(\mathbf{s}_t))^2$.
Given a sampled action $a_t$ and the estimate of cumulative rewards $A_t$, the optimization process based on the above expression can be imagined as that of shifting $\mu_{\,\!_\theta}(\mathbf{s}_t)$ towards the direction of $a_t$ if $A_t$ is higher than the expectation, or to the opposite direction if $A_t$ is smaller. 
Such an approach, though, can easily converge to a suboptimal solution, if, for example, the reward landscape has a basis between the current location of $\mu_{\,\!_\theta}(\mathbf{s}_t)$ and the optimal solution, or hard to be optimized if the reward landscape is symmetric around $\mu_{\,\!_\theta}(\mathbf{s}_t)$. 
These issues arise due to the fact that the Gaussian distribution is inherently unimodal, while the reward landscape could be multi-modal~\cite{softq}. 
We refer to Appendix~\ref{app:multimodal} for further discussion about the limitations of unimodal Gaussian policies and the value of expressive multimodal policies.
Indeed, \citet{hamalainen2020visualizing} showed that the reward landscape for high-dimensional character control tasks typically has a complex shape and is often multimodal.
This means that Gaussian policies may face difficulties during optimization for character control.

\section{Particle Filtering Policy Network}\label{sec:pfpn}

In this section, we describe our Particle Filtering Policy Network (PFPN) that addresses the unimodality issues from which typical Gaussian-based policy networks suffer. 
Our approach represents the action policy as a mixture distribution obtained by adaptively discretizing the action space using state-independent particles. The policy network,
instead of directly generating actions, it is tasked with choosing particles, while the final actions are obtained by sampling from the selected particles.

\subsection{Particle-Based Action Policy}\label{sec:policy}
We define $\mathcal{P} := \{\langle \mu_{i,k},w_{i,k}(\mathbf{s}_t\vert\theta) \rangle \vert i=1,\ldots,n; k=1,\ldots,m\}$ as a weighted set of particles for continuous control problems having an $m$-dimensional action space and $n$ particles distributed on each action action space.  
Here, $\mu_{i,k}$ represents an atomic action location on the $k$-th dimension of the action space, and $w_{i,k}(\mathbf{s}_t\vert\theta)$
denotes the associated weight generated by the policy network with parameters $\theta$ given the input state $\mathbf{s}_t$.
Let $p_{i,k}(a_{i,k}|\mu_{i,k}, \xi_{i,k}) $ denote the probability density function of the distribution defined by the location $\mu_{i,k}$ and a stochastic process $\xi_{i,k}$ for sampling. 
Given $\mathcal{P}$, we define the action policy as factorized across the action dimensions:
\begin{equation}\label{eq:particle_policy}
    \pi_{\theta}^{\mathcal{P}}(\mathbf{a}_t \vert \mathbf{s}_t) = \prod_k \sum_i w_{i,k}(\mathbf{s}_t |\theta)\ p_{i,k}(a_{t,k} |\, \mu_{i,k},\,\xi_{i,k}), 
\end{equation}
where $\mathbf{a}_t=\{a_{t,1}, \cdots, a_{t,m}\}$,  $a_{t,k}$ is the sampled action at the time step $t$ for the action dimension $k$, and $w_{i,k}(\cdot \vert \theta)$ 
is obtained by applying a softmax operation to the output of the policy network for the $k$-th dimension and satisfies $\sum_i w_{i,k} = 1$. 
The state-independent parameter set, $\{\mu_{i,k}\}$, gives us an adaptive discretization scheme that can be optimized during training.   
The noise parameter, $\xi_{i,k}$, provides a way to generate stochastic actions during policy training.
In our implementation, we choose Gaussian distributions with a standard deviation of $\xi_{i, k}$ for action sampling during training and then each particle can be regarded as a state-independent Gaussian of $\mathcal{N}(\mu_{i,k}, \xi_{i,k}^2)$.
Without loss of generality, we define the parameters of a particle as $\phi_{i,k} = [\mu_{i,k}, \xi_{i,k}]$ for the following discussion.

While the softmax operation gives us a categorical distribution defined by $w_{\cdot,k}(\mathbf{s}_t \vert \theta)$, the nature of the policy for each dimension is a mixture distribution with state-independent components defined by $\phi_{i,k}$.
The number of output neurons in PFPN increases linearly as the number of action dimensions increases, and thus makes it suitable for high-dimensional control problems.
Drawing samples from the mixture distribution can be done in two steps. first, based on the weights $w_{\cdot,k}(\mathbf{s}_t \vert \theta)$, we perform sampling on the categorical distribution to choose a particle $j_k$ for each dimension $k$, i.e. 
\begin{equation}
    j_k(\mathbf{s}_t) \sim P(\cdot \vert w_{\cdot,k}(\mathbf{s}_t)).
\end{equation}
Then, we can draw actions from the components represented by the chosen particles with noise as
\begin{equation}
    a_{t,k} \sim p_{j_k(\mathbf{s}_t)}(\cdot \vert \phi_{j_k(\mathbf{s}_t)}).
\end{equation}

\subsection{Training}
The proposed particle-based policy distribution is general and can be applied directly to any algorithm using the policy gradient method with Equation~\ref{eq:policy_gradient}. 
To initialize the training, due to lack of prior knowledge, the particles can be distributed uniformly along the action dimensions with a standard deviation covering the gap between two successive particles. 
With no loss of generality, let us consider below only one action dimension and drop the subscript $k$. 
Then, at every training step, each particle $i$ will move 
along its action dimension and be updated by
\begin{equation}\label{eq:particle_update}
    \nabla J(\phi_{i}) = \mathbb{E}\left[\sum_t c_{t} w_{i}(s_t\vert\theta)\nabla_{\phi_i}p_{i}(a_{t}\vert \phi_i)\vert \mathbf{s}_t\right]
\end{equation}
where $a_t \sim \pi_\theta^\mathcal{P}(\cdot \vert \mathbf{s}_t)$ is the  action chosen during sampling,
and $c_{t} = \dfrac{A_t}{\sum_j w_{j}(\mathbf{s}_t\vert\theta)p_{j}(a_{t}\vert \phi_j)}$
is a coefficient shared by all particles on the same action dimension.
Our approach focuses only on the action policy representation in general policy gradient methods. 
The estimation of $A_t$ can be chosen as required by the underlying policy gradient method, e.g. the generalized advantage estimator~\cite{schulman2015high} in PPO/DPPO. 
Similarly, for the update of the policy neural network, we have
\begin{equation}\label{eq:policy_update}
    \nabla J(\theta) = \mathbb{E}\left[\sum_t c_{t} p_{i}(a_{t}\vert\phi_i) \nabla_\theta w_{i}(\mathbf{s}_t \vert \theta) \vert \mathbf{s}_t\right].
\end{equation}

From the above equations, although sampling is performed on only one particle for each given dimension, all of that dimension's particles will be updated during each training iteration to move towards or away from the location of $a_{t}$ according to $A_t$.
The amount of the update, however, is regulated by the state-dependent weight $w_{i}(\mathbf{s}_t \vert \theta)$:
particles that have small probabilities to be chosen for a given state $\mathbf{s}_t$ will be considered as uninteresting and be updated with a smaller step size or not be updated at all. 
On the other hand, the update of weights is limited by the distance between a particle and the sampled action: particles too far away from the sampled action would be considered as insignificant to merit any weight gain or loss. 
In summary, particles can converge to different optimal locations near them during training
and be distributed multimodally according to the reward landscape defined by $A_t$, rather than collapsing to a unimodal, Gaussian-like distribution.

\subsection{Resampling}\label{sec:resampling}

Similar to traditional particle filtering approaches, our approach would encounter the problem of degeneracy~\cite{kong1994sequential}. 
During training, a particle placed near a location at which sampling gives a low $A_t$ value would achieve a  weight decrease. Once the associated weight reaches near zero, the particle will not be updated anymore (cf. Equation~\ref{eq:particle_update}) and become `dead'. 
We adapt the idea of resampling from the particle filtering literature~\cite{doucet2001introduction} to perform resampling for dead particles and reactivate them by duplicating alive target particles. 

A particle is considered dead if its maximum weight over all possible states is too small, i.e.
\begin{equation}
    \max_{\mathbf{s}_t} w_{i}(\mathbf{s}_t \vert \theta) < \epsilon
\end{equation}
where $\epsilon$ is a small positive threshold number. In practice, we cannot 
check $w_{i}(s_t \vert \theta)$ for all possible states, but can keep tracking it during sampling based on the observed states collected in the last batch of environment steps. During resampling, a target particle $\tau_i$ is drawn for each dead particle $i$ independently.
We consider two resampling strategies: 
(1) the unweighted resampling strategy that picks 
a $\tau_i$ randomly from all alive particles; and 
(2) the weighted resampling strategy that draws a target $\tau_i$  from the categorical distribution obtained from the average weight of each particle over the observed samples, i.e.
\begin{equation}
    \tau_i \sim P\left(\cdot \vert \mathbb{E}_{\mathbf{s}_t}\left[w_{k}(\mathbf{s}_t\vert\theta)\right], k=1,2,\cdots\right).
\end{equation}
Both of these two resampling strategies are stochastic. 
We, by default, use the weighted resampling, which considers the importance of the target particle candidates leading to a more stable learning performance, as we will show empirically in Section~\ref{sec:ablation}.

\begin{algorithm}[t]
\SetAlgoLined\small
Initialize the neural network parameter $\theta$ and learning rate $\alpha$\;
initialize particle parameters $\phi_i$ to uniformly distribute particles on each action dimension\;
initialize the threshold $\epsilon$ to detect dead particles\;
initialize the value of interval $n$ to perform resampling.
 
 \While{training does not converge}{
      \For{each environment step}{
            // \textit{Record the weight while sampling.}\\
            $a_t \sim \pi_{\theta,\mathcal{P}}(\cdot \vert s_t)$;
            $\mathcal{W}_i \gets \mathcal{W}_i \cup \{w_{i}(s_t\vert\theta)\}$
      }
      \For{each training step}{
            // \textit{Update parameters using SGD method.}\\
            $\phi_i \gets \phi_i + \alpha \nabla J(\phi_{i})$\quad// \textit{Equation~\ref{eq:particle_update}}\\
            $\theta \gets \theta + \alpha \nabla J(\theta)$\quad// \textit{Equation~\ref{eq:policy_update}}
      }  
      \For{every $n$ environment steps}{
            // \textit{Detect dead particles and set up target ones.}\\
            \For{each particle $i$}{
                \If{$\max_{w_i\in\mathcal{W}_i} w_i < \epsilon$}{
                    $\tau_i \sim P\left(\cdot \vert \mathbb{E}\left[w_{k}\vert w_k \in \mathcal{W}_k \right], k=1,2,\cdots\right)$ \\
                    $\mathcal{T} \gets \mathcal{T} \cup \{\tau_i\}$; $\mathcal{D}_{\tau_i} \gets \mathcal{D}_{\tau_i} \cup \{i\}$
                }
            }
            // \textit{Resampling.}\\
            \For{each target particle $\tau \in \mathcal{T}$}{
                \For{each dead particle $i \in \mathcal{D}_\tau$}{
                    // \textit{Duplicate particles.}\\
                    $\phi_i \gets \phi_\tau$ with $\mu_i \gets \mu_\tau + \varepsilon_i$ \\
                    // \textit{Duplicate parameters of the last layer in the policy network.} \\
                    $\boldsymbol{\omega}_i \gets \boldsymbol{\omega}_\tau$; $b_{i} \gets b_{\tau} - \log(\vert\mathcal{D}_\tau\vert + 1)$
                }
                $b_{\tau} \gets b_{\tau} - \log(\vert\mathcal{D}_\tau\vert + 1)$;
                $\mathcal{D}_\tau \gets \emptyset$
            }
            $\mathcal{T} \gets \emptyset$; $\mathcal{W}_i \gets \emptyset$ 
      }  
 }
 \caption{Policy Gradient Method using PFPN}
 \label{alg}
\end{algorithm}

\begin{figure}[t]
    \centering
    \includegraphics[width=.9\linewidth]{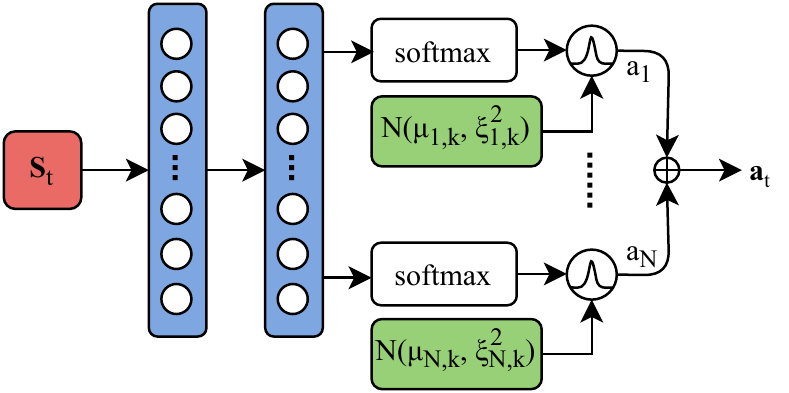}
    \caption{PFPN architecture with a $N$-dimension action space in our experiment. Each particle represents a state-independent Gaussian distribution of $\mathcal{N}(\mu_{i,k}, \xi^2_{i,k})$ where $k=1, \cdots, 35$ for 35 particles on each action dimension and $i=1,\cdots,N$. $\oplus$ denotes the concatenation operator.}
    \label{fig:network}
\end{figure}

\begin{theorem*}
Let $\mathcal{D}_{\tau}$ be a set of dead particles sharing the same target particle $\tau$. 
Let also the logits for the weight of each particle $k$ be generated by a fully-connected layer with parameters $\boldsymbol\omega_k$ for the weight and $b_k$ for the bias.
The policy  $\pi_\theta^{\mathcal{P}}(a_t \vert \mathbf{s}_t)$ is guaranteed to remain unchanged after resampling via duplicating $\phi_i \gets \phi_{\tau}, \forall i \in D_{\tau}$, if
the weight and bias used to generate the unnormalized logits of the target particle are shared with those of the dead one as follows:
\begin{equation}
    \boldsymbol\omega_i \gets \boldsymbol\omega_{\tau}; \quad b_i, b_{\tau} \gets b_{\tau} - \log\left(\vert\mathcal{D}_{\tau}\vert + 1\right).
\end{equation}
\end{theorem*}
\begin{proof}
See Appendix~\ref{app:resample} for the inference.
\end{proof}

The theorem guarantees the correctness of our resampling process as it keep the action policy $\pi_\theta^{\mathcal{P}}(a_t \vert \mathbf{s}_t)$ identical before and after resampling.
If, however, two particles are exactly the same after resampling, they will always be updated together at the same pace during training and lose diversity. 
To address this issue, we add some regularization noise to the mean value when performing resampling,
i.e. $\mu_i \gets \mu_{\tau} + \varepsilon_i$, 
where $\varepsilon_i$ is a small random number to prevent $\mu_{i}$ from being too close to its target $\mu_{\tau}$. We refer to Algorithm~\ref{alg} for the outline of our proposed PFPN approach.

\begin{figure*}[ht]
    \centering
    \includegraphics[width=.9\linewidth]{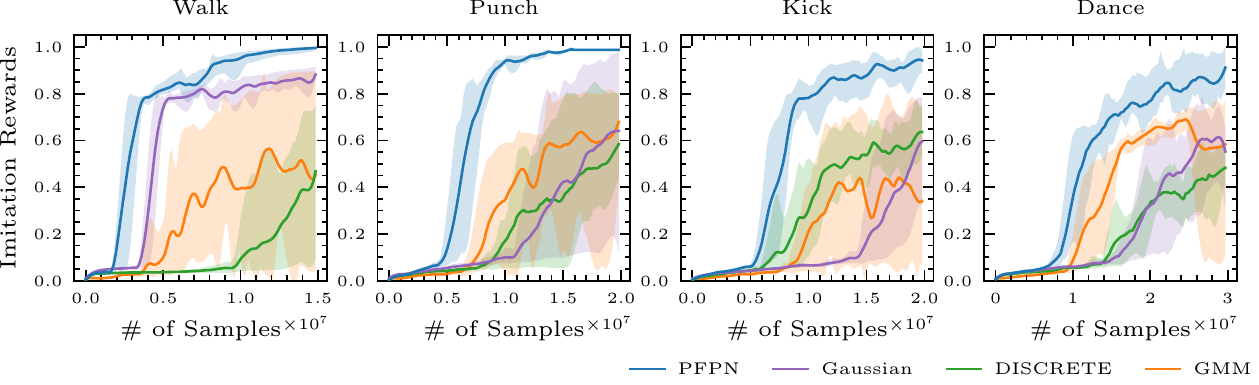}
    \caption{Learning curves of PFPN compared to other baselines using DPPO algorithm on humanoid character control tasks. Solid lines report the average and shaded regions are the minimum and maximum imitation rewards achieved with different random seeds during training.}
    \label{fig:main}
\end{figure*}

\begin{figure}[ht]
    \centering
    \includegraphics[width=0.85\linewidth]{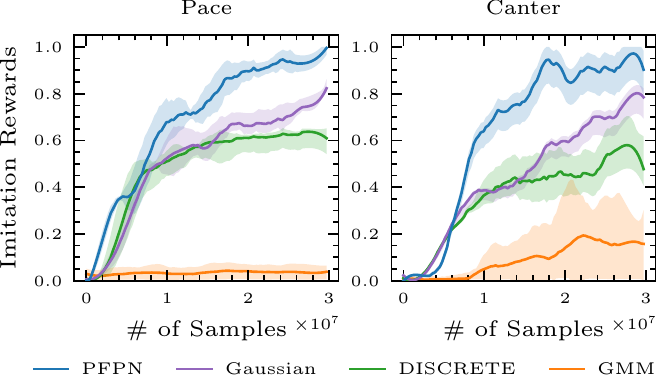}
    \caption{Learning performance of baselines using DPPO for dog character control.
    }
    \label{fig:main_dog}
\end{figure}

\subsection{Action-Value Based Optimization}
Algorithm~\ref{alg} can be applied on general policy gradient algorithms, like PPO~\cite{schulman2017proximal} and A3C~\cite{mnih2016asynchronous}.
However,
we note that a number of algorithms, such as DDPG~\cite{lillicrap2015continuous}, SAC~\cite{haarnoja2018soft, haarnoja2018soft2} and their variants~\cite{softq,fujimoto2018addressing},
perform optimization by maximizing a soft state-action value $Q(\mathbf{s}_t, \mathbf{a}_t)$.
In this case,
the action policy is required to be reparameterizable such that the sampled action $\mathbf{a}_t$ can be rewritten as a function differentiable to the policy network parameter $\theta$,
and the optimization can be done through the gradient $\nabla_{\mathbf{a}_t} Q(\mathbf{s}_t, \mathbf{a}_t)\nabla_\theta\mathbf{a}_t$.

Our two-step sampling method for PFPN described in Section~\ref{sec:policy} is non-reparameterizable, because of the standard way of sampling from the categorical distribution through which Gaussians are mixed. 
To address this issue and enable the proposed action policy applicable in state-action value based off-policy algorithms, 
we consider the concrete distribution~\cite{jang2016categorical, maddison2016concrete} that generates a reparameterized continuous approximation to a categorical distribution.
We refer to Appendix~\ref{app:reparameter} for the reprarameterization trick and further details on the application of PFPN to off-policy algorithms.


\section{Experiments}
The goal of our experiments is to evaluate whether PFPN can outperform the corresponding implementations with Gaussian policies in data-driven control tasks of physics-based character.
We focus on the two aspects of comparisons: (1) the imitation ability of the trained policies, and (2) the policy robustness facing external force perturbation. 
We also perform sensitivity analysis on the learning performance of PFPN with different number of particles and resampling strategies.

\begin{figure*}[ht]
    \centering
    \begin{subfigure}[b]{0.33\linewidth}
        \centering
        \includegraphics[width=\linewidth]{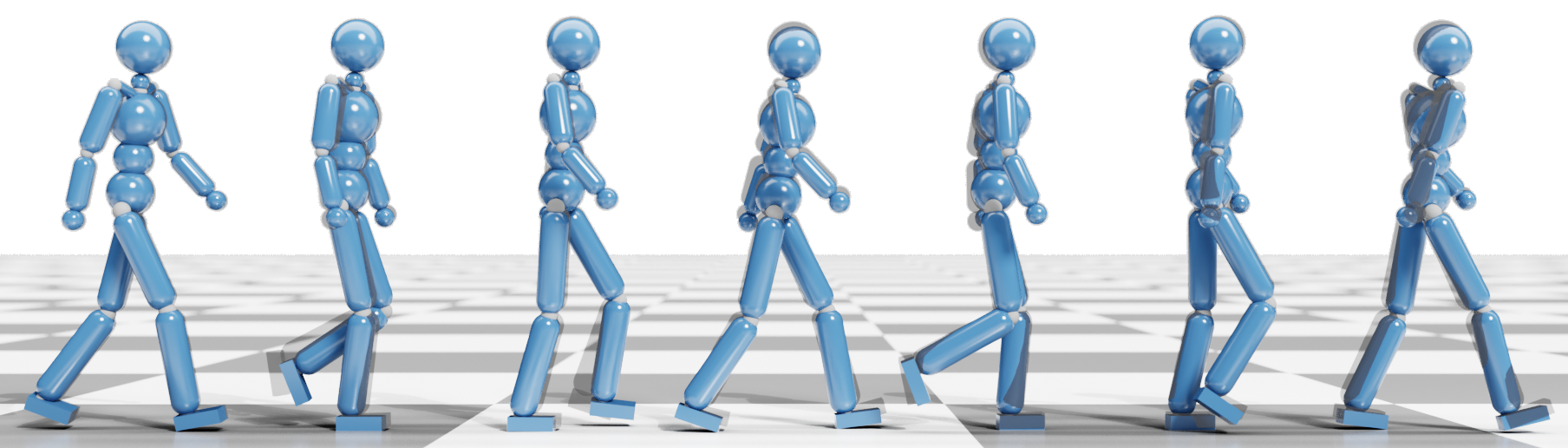}
        \includegraphics[width=\linewidth]{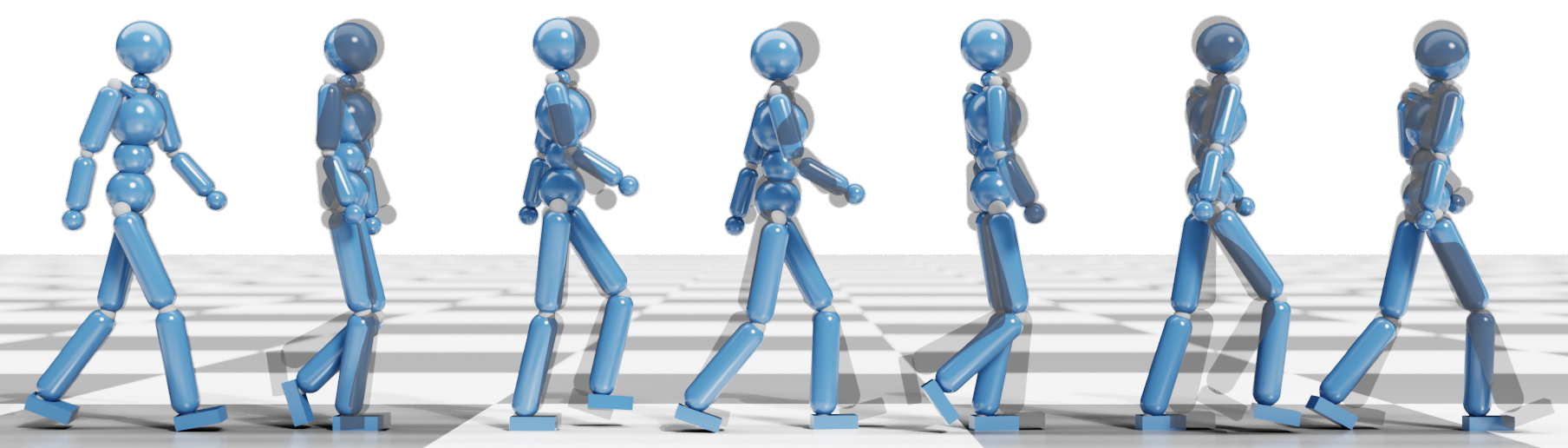}
        \caption{Walk}
    \end{subfigure}
    \begin{subfigure}[b]{0.33\linewidth}
        \centering
        \includegraphics[width=\linewidth]{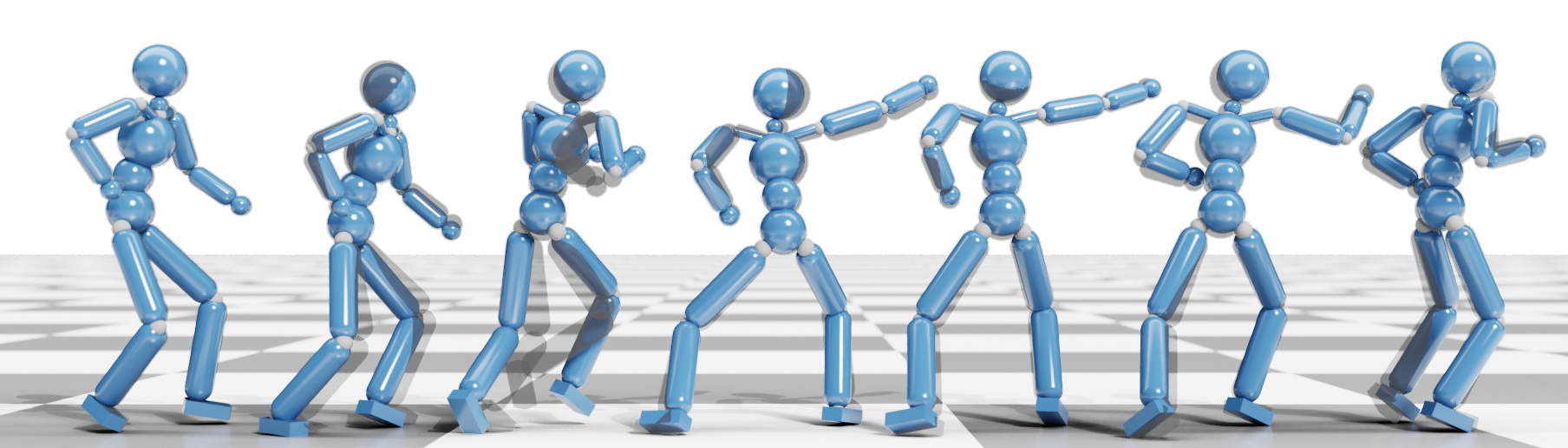}
        \includegraphics[width=\linewidth]{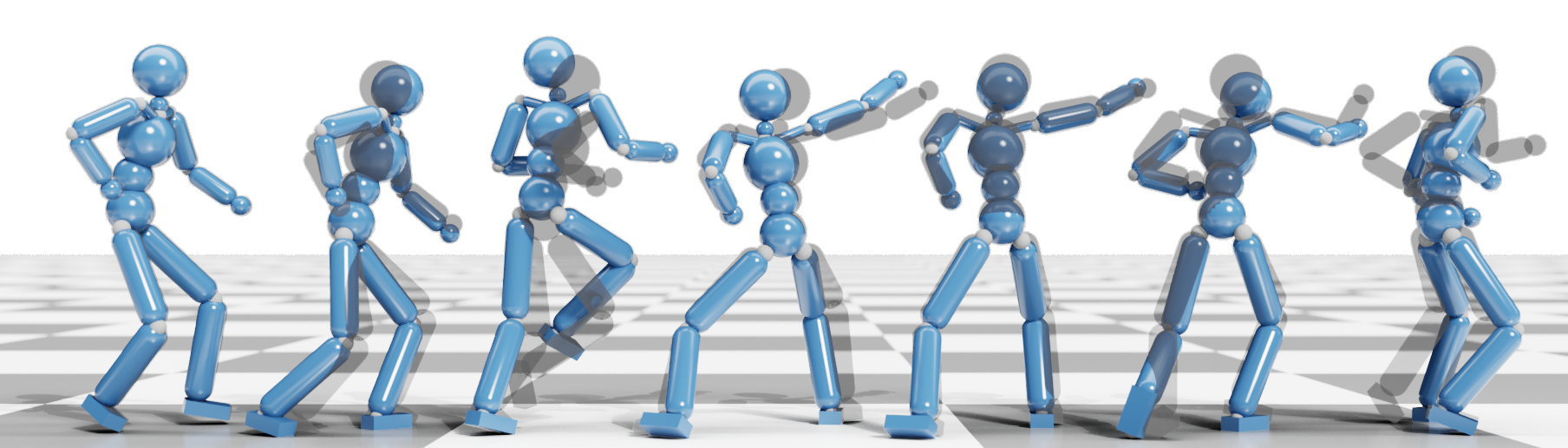}
        \caption{Punch}
    \end{subfigure}
    \begin{subfigure}[b]{0.33\linewidth}
        \centering
        \includegraphics[width=\linewidth]{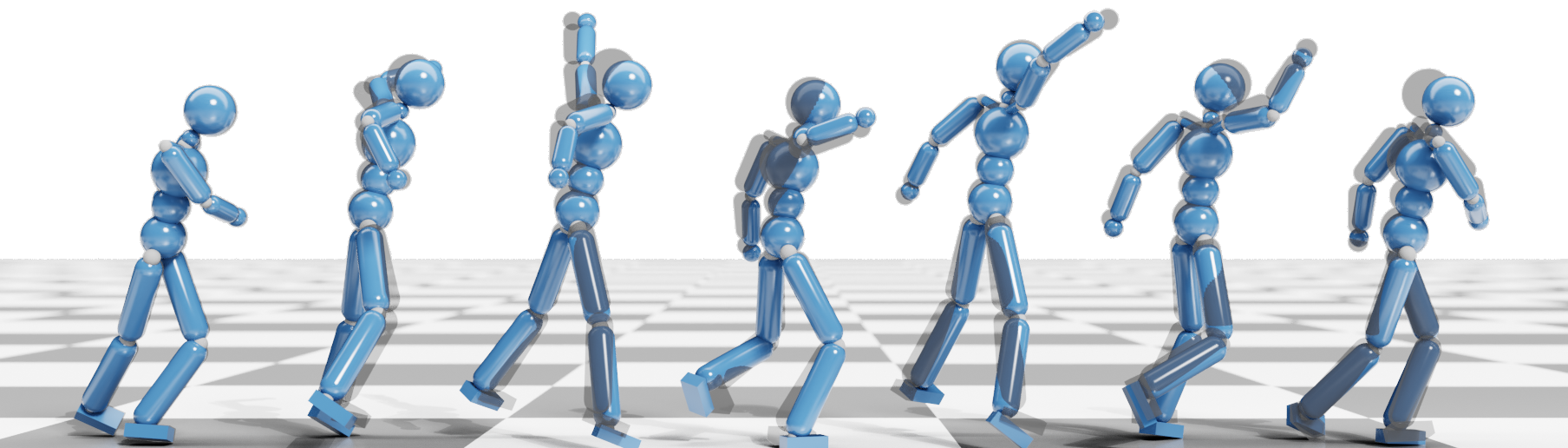}
        \includegraphics[width=\linewidth]{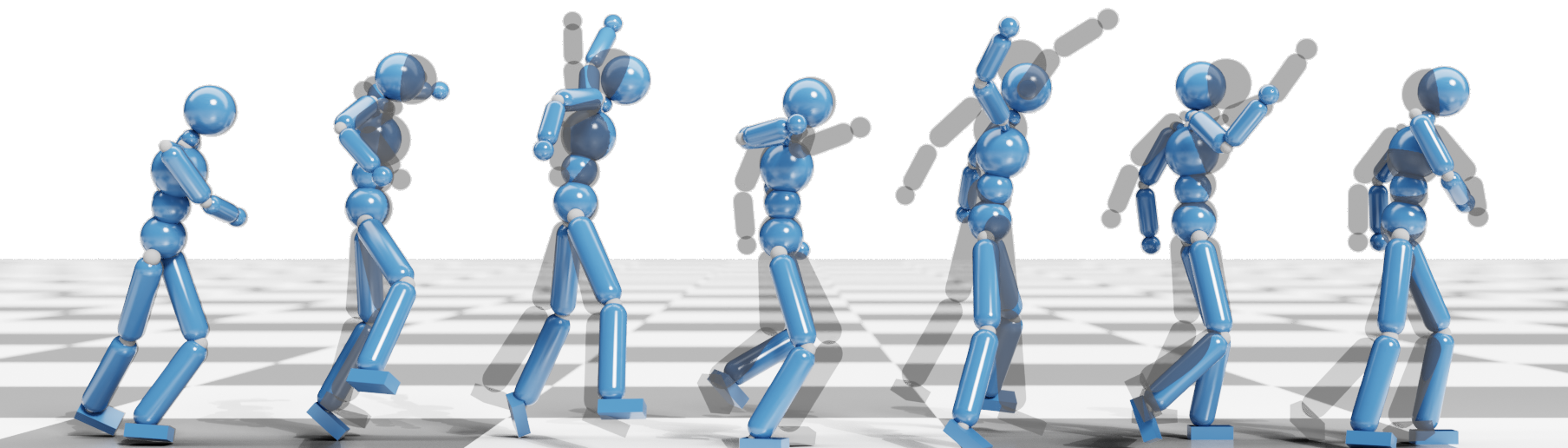}
        \caption{Dance}
    \end{subfigure}
    \caption{
    Qualitative comparisons of the motions generated by PFPN (top) and Gaussian (bottom) baselines.
    Shadow characters indicate the reference motion. 
    }
    \label{fig:motions}
\end{figure*}

\begin{figure*}[ht]
\centering
\includegraphics[width=.75\linewidth]{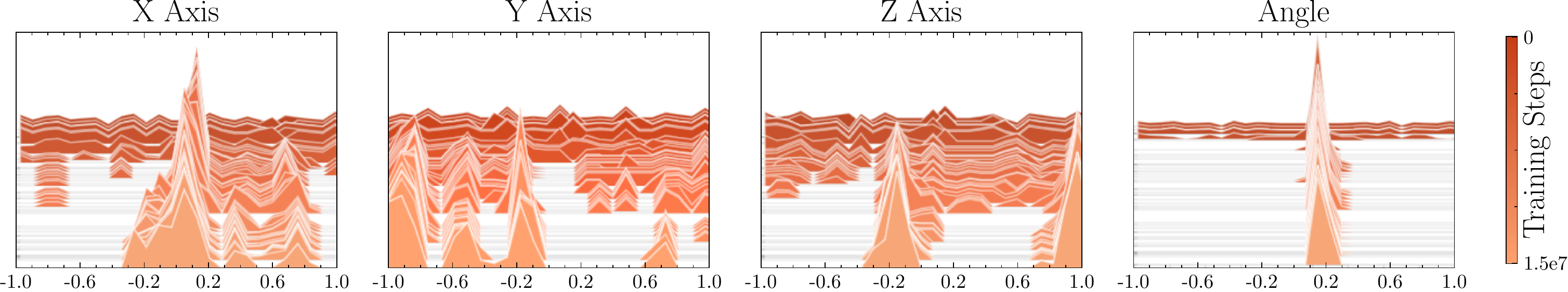}
\caption{Evolution of how particles along the four action dimensions of the right hip joint are distributed during training of the Walk task.
Each action dimension is normalized between -1 and 1.  
Particles are initially distributed uniformly along a dimension (dark colors) and their locations adaptively change as the
policy network is trained (light colors). The training steps are measured by the number of samples exploited during training.
}
\label{fig:particle_evolution}
\end{figure*}

\subsection{Setup}
We run benchmarks based on the DeepMimic framework~\cite{peng2018deepmimic}, which is the state-of-the-art DRL imitation learning framework for physics-based character control. 
The simulated character is controlled through stable proportional derivative controllers~\cite{tan2011stable} running with the forward dynamic simulation at 600 Hz, while the policy network provides the control signal at 30 Hz.
Following the settings of DeepMimic, we use the link position, orientation (in the unit of quaternion) and linear and angular velocity related to the root link position and heading direction, adding a phase variable to indicate the target pose implicitly at each time step and a variable indicating the root link height, as the observation space. 
We considered two articulated characters, a humanoid and a dog.  
The humanoid character has 8 spherical joints and 4 revolute joints, plus a root link and two end-effectors (hands) connected to the forearms with fixed joints, resulting in an observation space of $\mathbb{R}^{197}$ and action space of $\mathbb{R}^{36}$.
The dog character has 18 spherical joints and 4 revolute joints, which lead an observation space of $\mathbb{R}^{301}$ and action space of $\mathbb{R}^{76}$.
All simulations are run using PyBullet~\cite{coumans2021}.

PFPN focuses only on the distribution for action policy and can be applied with any general policy-gradient DRL algorithm.
In the following, we evaluate PFPN with policies trained using DPPO~\cite{heess2017emergence} with surrogate policy loss~\cite{schulman2017proximal} (We refer to Appendix~\ref{app:results} for results obtained with the A3C~\cite{mnih2016asynchronous} and IMPALA  algorithms~\cite{espeholt2018impala}, and to  Appendix~\ref{app:hyper} for all hyperparameters used).
In our DPPO implementation, policy and value networks have a similar structure of two hidden fully-connected (FC) layers with neurons of 1024 and 512 respectively, as shown in Figure~\ref{fig:network}.
The input state is normalized by moving average that is dynamically updated during training.
By default, we place 35 particles on each action dimension and use the set of particles as a mixture of Gaussians with state-dependent weights but state-independent components.

\subsection{Baseline Comparison}
In Figure~\ref{fig:main} and~\ref{fig:main_dog}, we compare PFPN to baselines using Gaussian distribution with state-dependent mean and standard deviation values on the humanoid and dog character control tasks respectively. 
The imitation performance is measured in the term of normalized cumulative rewards of evaluation rollouts during training.
We also compare PFPN to a fixed discretization scheme (DISCRETE) obtained by uniformly discretizing each action dimension into a fixed number of bins and sampling actions from a categorical distribution, and to policies using the distribution of a fully state-dependent Gaussian Mixture Model (GMM).
All baselines use the same network architecture with the same number of hidden neurons.
PFPN, DISCRETE and GMM also have the same number of atomic actions (35 particles/bins) at each action dimension. 
We train five trials of each baseline with different random seeds that are the same across PFPN and the corresponding implementations of other methods.
Evaluation was performed ten times every 1,000 training steps using deterministic actions. 

As it can be seen from the figures, PFPN outperforms other baselines in all of the tested tasks. 
Our particle-based scheme achieves better final performance and exhibits faster convergence while being more stable across multiple trials.
GMM uses Gaussian components with state-dependent mean and standard deviation values, and does not work better than Gaussian baselines.
DISCRETE discretizes the action space using the atomic actions, which are the same with the initial distribution of the particles employed by PFPN.
Though both DISCRETE and PFPN use the state-independent atomic action settings, PFPN 
optimizes the distribution of atomic actions represented by particles during training, and the resulting adaptive discretization scheme 
leads to higher asymptotic performance.
Intuitively, by introducing more atomic actions, DISCRETE and PFPN could achieve higher control capacity.
However, the more atomic actions employed the harder the optimization problem will be due to the increase of policy gradient variance~\cite{tang2019discretizing} (see Appendix~\ref{app:grad_var} for theoretical analysis).
In theory, GMM is a more general case of PFPN without resampling. However, we found that GMM does not usually work quite well and can perform even worse than Gaussians. This is consistent with the results reported by \citet{tang2018boosting} for torque-based locomotion control tasks. A possible reason is that GMM needs a much larger policy network. This may pose a challenge to the optimization. Another issue of GMMs observed during our experiments is that the Gaussian components are easy to collapse together and lose the advantage of multimodality.

\begin{figure}[t]
    \centering
    \includegraphics[width=\linewidth]{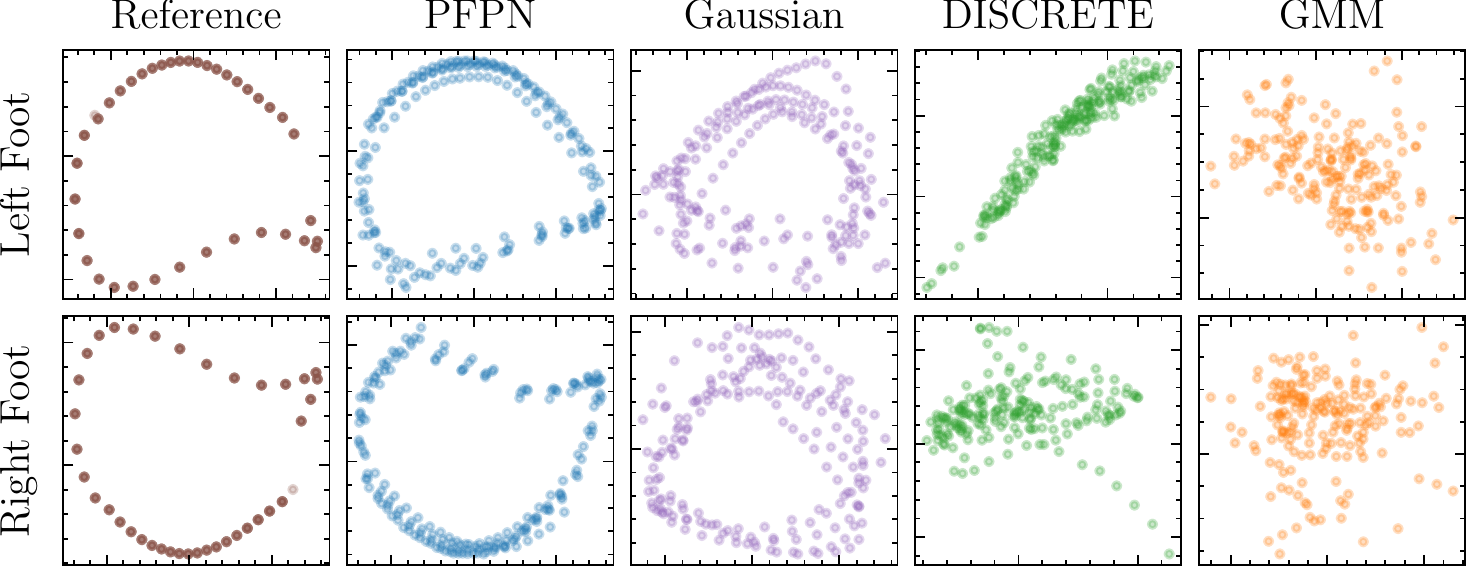}
    \caption{PCA embedding of the character's foot trajectories related to the base link position during ten gait cycles of walking. The reference one is one-gait walking motion obtained through motion capture.}
    \label{fig:motion_walk_pca}
\end{figure}

To gain a better understanding of the advantages of using PFPN
for learning motor tasks, Figure~\ref{fig:particle_evolution} shows how particles evolve during training for one of the humanoid's joints in the ``Walk'' task.
We can see that each of the final active action spaces follows a multimodal distribution that covers only some small parts of the entire action space.
PFPN optimizes the placement of atomic actions, providing an effective discretization scheme that reaches better performance compared to other baselines.

\subsection{Motion Quality}
In Figure~\ref{fig:motion_walk_pca}, we compare the foot trajectories of each baseline during ten gait cycles in ``Walk'' tasks.
From the figure, GMM generates jittery motions with unstable foot trajectories.
DISCRETE can provide relatively stable trajectories, in which, however, the character moves always towards an inclined direction.
Gaussian gives gaits with a cyclic pattern but not quite stable.
PFPN generate stable gaits with a clear cyclic pattern closely following the reference motion.
In Figure~\ref{fig:motions}, we qualitatively compare the motions generated by PFPN and Gaussian baselines.
As can be seen, the motion generated by PFPN follows the reference motion (shadow character) closely.
Gaussian, though providing a human-like walking motion, exhibits visual artifacts in the other, more complex tasks, 
e.g. foot sliding and character drifting during dancing.
Similar conclusions can be drawn for the dog character. 
When a fixed number of samples are exploited for training as reported in Figure~\ref{fig:main_dog},  
PFPN baselines can track the reference motion closely while Gaussian baselines perform worse, with the dog, for example, having evident jerky movements during pacing. 

In addition, we note that characters trained with Gaussian-based policies often lack the grace seen in living beings as compared to PFPN-trained characters that exhibit behavior more close to the one seen in the reference data. For example, while the dog is able to achieve a relatively high reward according to DeepMimic's tracking reward function (Figure~\ref{fig:main_dog}), its tail moves in an unnatural way during canter. Similarly, while this is more subtle, the Gaussian-based humanoid character stamps its feet on the ground while walking, resulting in a heavy-footed gait as compared to the graceful gait of the PFPN-based humanoid. 
We refer to the supplemental videos for more details on the visual quality of the trained controllers.

\subsection{Robustness}
\begin{figure}[t]
    \centering
    \includegraphics[width=0.9\linewidth]{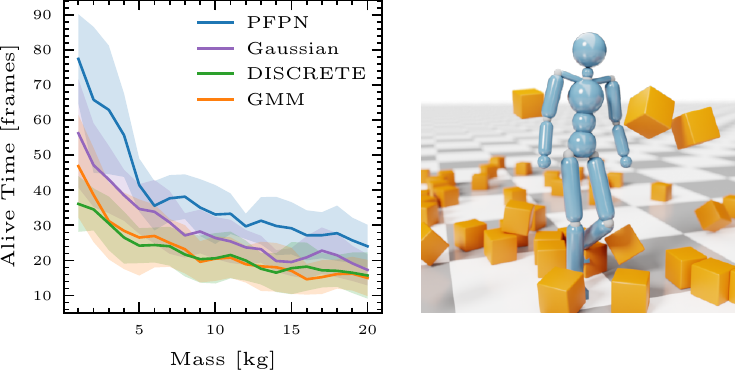}
    \caption{Policy robustness of ``Walk'' motion with cube projectiles of varying mass. Solid lines report the average performance over ten trials and shaded regions indicate the standard deviation. All policies are obtained using DPPO algorithms without projectile training. Right: a character under projectile testing. }
    \label{fig:robust}
\end{figure}

We evaluate policy robustness through projectile testing and also external force perturbations. During projectile testing, the character is controlled by the ``walk'' policy, and a cube projectile with side length of 0.2m and initial velocity of 0.2m/s is cast towards the torso of the character.
Figure~\ref{fig:robust} reports the number of frames that a policy can control the character before falling down on the ground while varying the mass of the cube. 
In Table~\ref{tab:force_perb}, we also report the minimal force needed to push the  
character down in ``walk'' and ``punch'' tasks for the humanoid, and in ``pace'' and ``canter'' tasks for the dog character. All experiments were performed after training using deterministic actions. 
It is evident that the character controlled by PFPN
are more robust to external disturbances and can sustain much higher forces than other baselines. 

\begin{table}[t]
    \caption{Minimal forward and sideways push needed to make the character fall down. Push force is measured in Newtons ($N$) and applied on the chest of the character for 0.1s.
    The DISCRETE and GMM results are skipped for the dog character, as the respective trained controllers are unable to make the character walk even in the absence of external disturbances. 
    }
    \centering\small
    \begin{tabular}{cccccc}
         Task & Force Direction & PFPN & Gaussian & DISCRETE & GMM \\
         \toprule
         \multicolumn{2}{l}{\textit{Humanoid}} &&&& \\
         \midrule
        \multirow{2}{*}{Walk}
        & Forward & $\mathbf{588}$ & $512$ & $340$ & $373$\\
        & Sideway & $\mathbf{602}$ & $560$ & $420$ & $417$\\
        \midrule
        \multirow{2}{*}{Punch}
        & Forward & $\mathbf{1156}$ & $720$ & $480$ & $721$\\
        & Sideway & $\mathbf{896}$ & $748$ & $576$ & $732$\\
        \toprule
        \multicolumn{2}{l}{\textit{Dog}} &&&& \\
        \midrule
        \multirow{2}{*}{Pace}
        & Forward & $\mathbf{735}$ & $672$ & - & -\\
        & Sideway & $\mathbf{412}$ & $276$ & - & -\\
        \midrule
        \multirow{2}{*}{Canter}
        & Forward & $\mathbf{418}$ & $404$ & - & -\\
        & Sideway & $\mathbf{344}$ & $274$ & - & -\\
        \bottomrule
    \end{tabular}
    \label{tab:force_perb}
\end{table}

\begin{figure}[t]
    \centering
    \begin{subfigure}[b]{\linewidth}
        \centering
        \includegraphics[width=\linewidth]{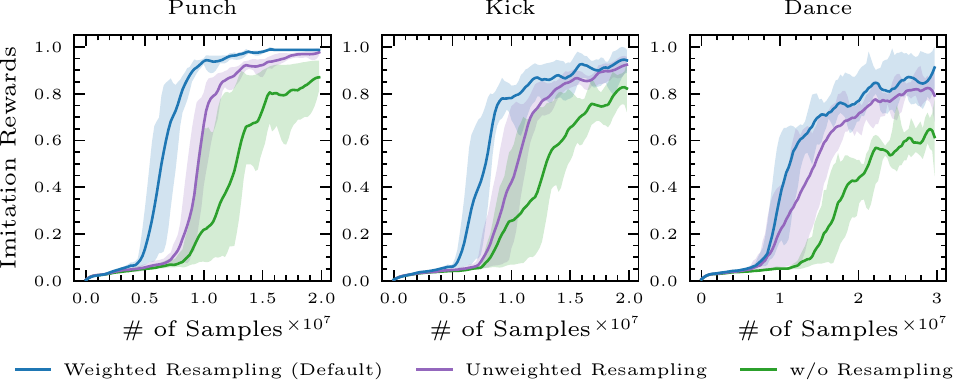}
        \vspace{-0.4cm}
        \caption{Learning performance of PFPN with 35 particles on each action dimension but different resampling strategies.}
        \label{fig:resampling}
    \end{subfigure}
    
    \begin{subfigure}[b]{\linewidth}
        \centering
        \includegraphics[width=\linewidth]{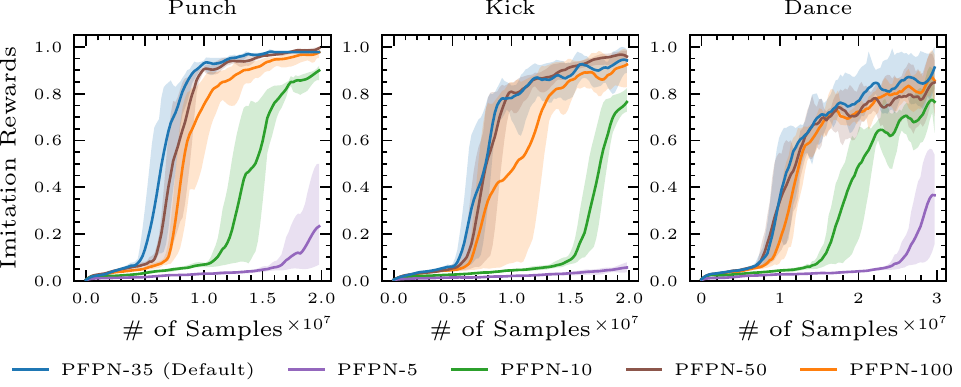}
        \vspace{-0.4cm}
        \caption{Comparison of PFPN using 35 particles per action dimension (PFPN-35) to that using 5, 10, 50 and 100 particles.}
        \label{fig:n_particles}
    \end{subfigure}
    \caption{Sensitivity of PFPN to different resampling strategies and the number of particles.}
    \label{fig:sensitivity}
\end{figure}

\subsection{Ablation Study}\label{sec:ablation}
\paragraph{Resampling Strategy.}
In Figure~\ref{fig:resampling}, 
we compare PFPN with default, weighted resampling  to PFPN with unweighted resampling (see Section~\ref{sec:resampling}), and to PFPN without any resampling. 
The weighted resampling strategy draws targets for dead particles according to the weights of the  remaining, alive ones. 
The unweighted resampling strategy draws targets uniformly from alive particles. 
It can be seen that both weighted and unweighted resampling could help improve the training performance significantly.
However, unweighted resampling could lead to high variance by introducing too much uncertainty, since it could reactivate dead particles and place them in suboptimal locations with higher probability, compared to the weighted resampling. 
After resampling, even though the particles would be optimized or resampled once more if they are placed in bad locations, this could make the training process converge slower, as shown in the figure.

\paragraph{Number of Particles.}
Since the particle configuration in PFPN is state-independent, it needs a sufficient number of particles to meet the fine control demand.
Intuitively, employing more particles will increase the resolution of the action space, and thus increase the control capacity and make fine control more possible.
However, in Appendix~\ref{app:grad_var}, we prove that due to the variance of policy gradient increasing as the number of particles increases, the more particles employed, the harder the optimization would be.
Therefore, it may negatively influence the performance to employ too many particles.
This conclusion is consistent with the results  shown in Figure~\ref{fig:n_particles}.
As it can be seen, PFPN with 5 and 10 particles per action dimension performs badly;
though using 50 or 100 particles per action dimension slows down the convergence speed a little bit, our approach is not sensitive in terms of the final learning performance when a relative large number of particles are employed.


\begin{figure}[t]
    \centering
    \includegraphics[width=\linewidth]{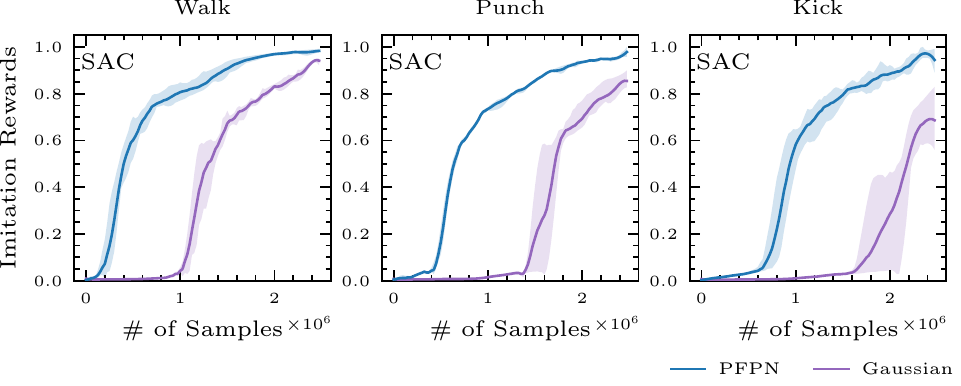}
    \caption{Learning performance of SAC with PFPN and Gaussian baselines.}
    \label{fig:sac}
\end{figure}
\subsection{PFPN Results with SAC}\label{sec:sac}
In this section, we highlight PFPN's performance in  state-of-the-art off-policy DRL algorithm of SAC~\cite{haarnoja2018soft2}.
As shown in Figure~\ref{fig:sac}, PFPN outperforms Gaussian baselines with faster convergence speed in all the tested tasks using SAC.
While SAC has been successfully explored for continuous control problems in the machine learning community,
DPPO is still the most commonly used DRL algorithm for training physics-based character controllers in the animation field.
During experiments, we found that PFPN with SAC is more stable in terms of learning performance and more sampling efficient compared to DPPO implementations as reported in Figure~\ref{fig:main}.
We refer to the supplementary video for
comparisons between SAC and DPPO.
Overall, PFPN with SAC needs only 2.5-million samples to 
train high-quality humanoid controllers in DeepMimic tasks as compared to DPPO that typically requires around 20 millions or even more samples to achieve the similar performance. 

\section{Discussion and Future Work}
We present PFPN as a general framework for systematic exploration of high-dimensional action spaces during training of physics-based character controllers. 
Our approach uses a mixture of state-independent Gaussians represented by a set of weighted particles to track the action policy, as opposite to the multivariate Gaussian that is typically used as the policy distribution in the tasks of physics-based character control.
We show that our method performs better than Gaussian baselines in various imitation learning tasks leading to  faster learning, higher motion quality and more robustness to external perturbation.

In the experimental section, we showed applications of PFPN to the PPO algorithm. However, our approach is applicable to other common on-policy actor-critic policy gradient DRL algorithms as we show in Appendix~\ref{app:results} and off-policy methods such as SAC, as we discussed in Section~\ref{sec:sac}. 
Overall, PFPN does not change the underlying architecture or learning mechanism of the DRL algorithms. 
It is, therefore, complementary to other techniques which improve the policy expressivity given a base action distribution. 
For example, PFPN can serve as the action policy for each expert in the mixture of experts approach of~\cite{ScaDiver} or for primitive action learning~\cite{peng2019mcp};
or as the base distribution of normalizing flows for motion generation~\cite{henter2020moglow},
which we would like to investigate in future work.

As shown in Section~\ref{sec:ablation}, PFPN is not quite sensitive to the number of particles when more than enough are employed to track the action distribution. 
However, some fine-tuning may be needed to determine the minimal number of particles necessary to achieve high learning performance with fast convergence speed. 
In this work, we only considered short-term imitation learning tasks. Thus, further experiments are needed to test the performance of PFPN for tracking long-term motions with heterogeneous behaviors~\cite{bergamin2019drecon,ScaDiver}.
Currently, we track each action dimension independently.
Accounting for the synergy that exists between different joints has the potential to further improve performance and motion robustness, which opens another exciting avenue for future work. 

In any case, we believe that our particle-based action policies provide a great alternative to Gaussian-based actions policies, which have been the staple for DRL-based character control over the past few years.  Our work shows significant improvements upon the state-of-the-art in terms of motion quality, robustness to external perturbations, and training efficiency, and we hope that more animation researchers would take advantage of PFPN while training physics-based controllers for continuous control tasks.


\begin{acks}
This work was supported in part by the National Science Foundation under Grant No. IIS-2047632.
\end{acks}

\bibliographystyle{ACM-Reference-Format}
\bibliography{mig2021-ref}

\clearpage\newpage
\appendix

\section{Multi-modal Policy}\label{app:multimodal}
In this section, we show the multi-modal representation capacity of PFPN on a one-step bandit task.

\begin{figure}[ht]
\centering
\includegraphics[width=0.7\linewidth]{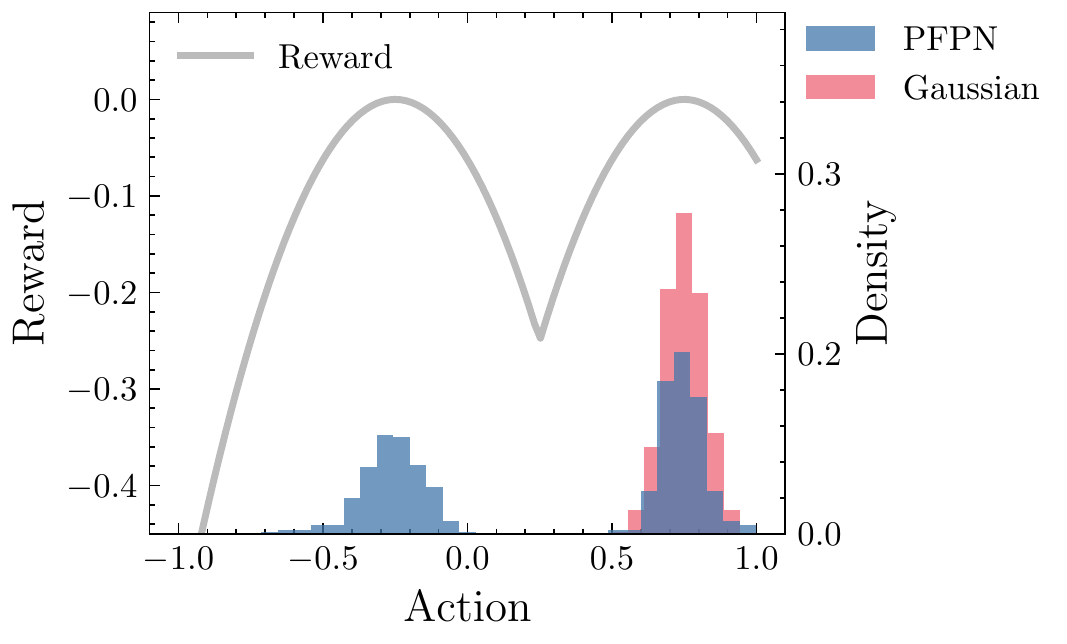}
\caption{One-step bandit task with asymmetric reward landscape. 
The reward landscape is defined as the gray line having two peaks asymmetrically at $-0.25$ and $0.75$.
The probability densities of stochastic action samples drawn from PFPN (blue)
and Gaussian policy (red) are counted after training with a fixed number of iterations.}
\label{fig:one_step_bandit_asy}
\end{figure}

This is a simple task with one dimension action space $\mathcal{A} = [-1, 1]$.
It has an asymmetric 2-peak reward landscape inversely proportional to the minimal distance to points $-0.25$ and $0.75$, as the gray line shown in Figure~\ref{fig:one_step_bandit_asy}.
The goal of this task is to find out the optimal points close to $-0.25$ and $0.75$.
In Figure~\ref{fig:one_step_bandit_asy}, we show the stochastic action sample distributions of PFPN and the naive Gaussian policy after training with the same number of iterations.
It is clear that PFPN captures the bi-modal distribution of the reward landscape, while the Gaussian policy gives an unimodal distribution capturing only one of reward peaks.

\section{Hyperparameters}\label{app:hyper}
\begin{table}[ht]
    \caption{Default hyperparameters in PFPN baselines.}
    \centering\small
    \begin{tabular}{ll}
        Parameter &  Value \\
        \toprule
        learning rate & $1\cdot10^{-4}$ \\
        resampling interval & 20 environment episodes  \\
        dead particle detection threshold ($\epsilon) $ & 0.0015  \\
        discount factor ($\gamma$) & $0.95$ \\
        clip range (DPPO) & 0.2  \\
        GAE discount factor (DPPO, A3C, $\lambda$) & 0.95  \\
        truncation level (IMPALA, $\bar{c}$, $\bar{\rho}$) & 1.0  \\
        coefficient of policy entropy loss term & \multirow{2}{*}{0.00025}\\
        \qquad\qquad\qquad\qquad\quad(A3C, IMPALA)&  \\
        reply buffer size (SAC) & $10^6$  \\
        \bottomrule
    \end{tabular}
    \label{tab:hyper}
\end{table}

Since it is infeasible to analytically evaluate the differential entropy of a mixture distribution without approximation, we use the entropy of the categorical distribution for A3C and IMPALA benchmarks, which employ differential entropy during policy optimization.

\section{Additional Results}\label{app:results}

\subsection{Time Complexity}
\begin{figure}[H]
    \centering
        \includegraphics[width=\linewidth]{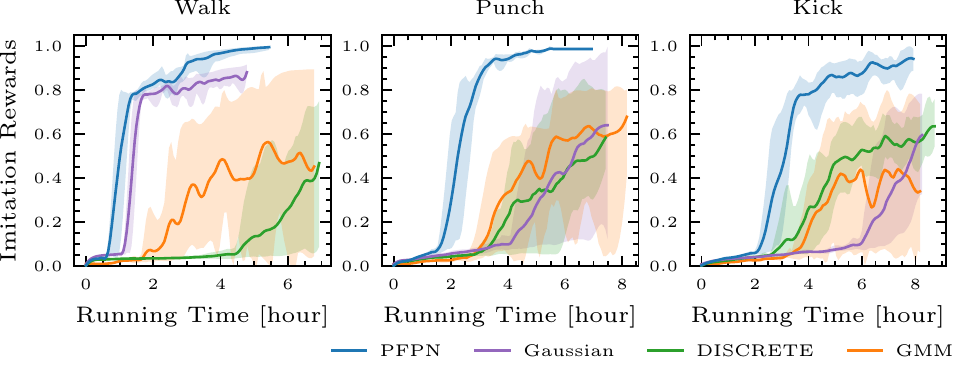}
    \caption{Learning performance as a function of the actual wall clock time using DPPO.}
\end{figure}
All policies were trained on a machine with Intel 6148G CPU and Nvidia V100 GPU.
Training stops when a fixed number of samples is collected as reported in Figure~\ref{fig:main}.
PFPN has a good time consumption performance compared to other baselines.
Though the action sampling and particles resampling processes would take extra time, PFPN performs better because its fast convergence avoids wasting time on environment reset when early termination occurs.

\subsection{Baselines}
\begin{figure}[ht]
    \centering
        \includegraphics[width=\linewidth]{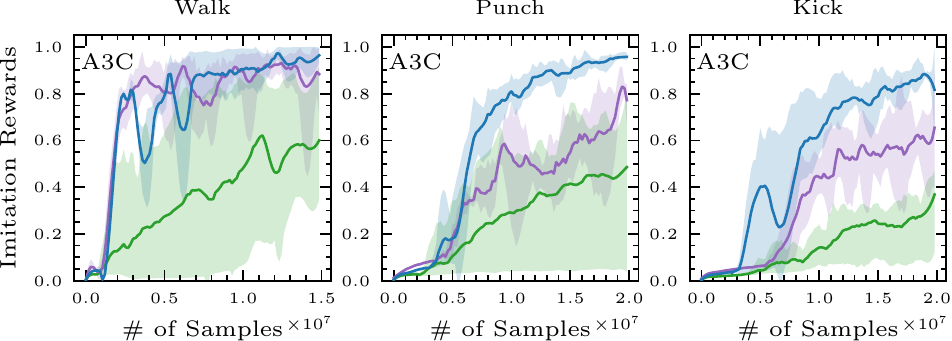}\\\vspace{0.2cm}
        
        \includegraphics[width=\linewidth]{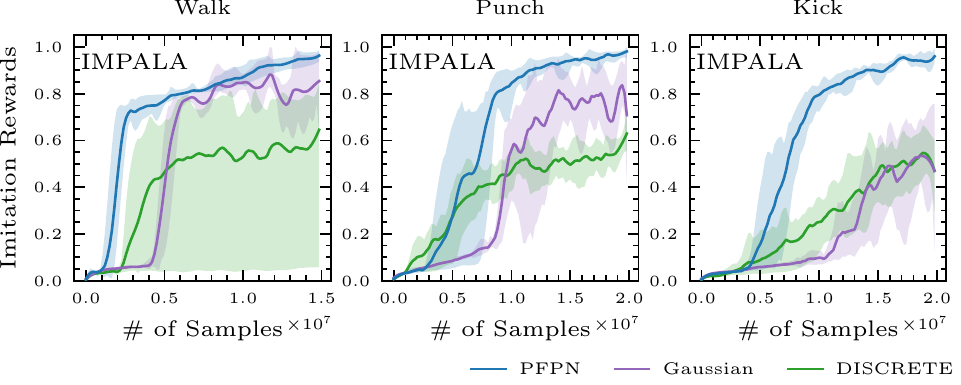}
    \caption{Additional baseline results using A3C and IMPALA.}
    \label{fig:add_baselines}
\end{figure}

\section{Policy Network Logits Correction during Resampling}\label{app:resample}

\begin{theorem*}
Let $\mathcal{D}_{\tau}$ be a set of dead particles sharing the same target particle $\tau$. 
Let also the logits for the weight of each particle $k$ be generated by a fully-connected layer with parameters $\boldsymbol\omega_k$ for the weight and $b_k$ for the bias.
The policy  $\pi_\theta^{\mathcal{P}}(a_t \vert \mathbf{s}_t)$ is guaranteed to remain unchanged after resampling via duplicating $\phi_i \gets \phi_{\tau}, \forall i \in D_{\tau}$, if
the weight and bias used to generate the unnormalized logits of the target particle are shared with those of the dead one as follows:
\begin{equation}
    \boldsymbol\omega_i \gets \boldsymbol\omega_{\tau}; \quad b_i, b_{\tau} \gets b_{\tau} - \log\left(\vert\mathcal{D}_{\tau}\vert + 1\right).
\end{equation}
\end{theorem*}
\begin{proof}
The weight for the $i$-th particle is achieved by softmax operation, which is applied to the unnormalized logits $L_i$, which is the direct output of the policy network:
\begin{equation}
    w_i(s_t) = \textsc{softmax}(L_i(s_t)) = \frac{e^{L_i(s_t)}}{\sum_k e^{L_k(s_t)}}.
\end{equation}

Resampling via duplicating makes dead particles become identical to their target particle. Namely, particles in $\mathcal{D}_{\tau}\cup\{\tau\}$ will share the same weights as well as the same value of logits, say $L_\tau^\prime$, after resampling. To ensure the policy identical before and after sampling, the following equation must be satisfied
\begin{equation}\label{eq:re_correct_1}
    \sum_k e^{L_k(s_t)} =\sum_{\mathcal{D}_{\tau}\cup\{\tau\}} e^{L_\tau^\prime(s_t)} + \sum_{k\not\in\mathcal{D}_{\tau}\cup\{\tau\}} e^{L_k(s_t)}
\end{equation}
where $L_k$ is the unnormalized logits for the $k$-th particle such that the weights for all particles who are not in $\mathcal{D}_{\tau}\cup\{\tau\}$ unchanged, while particles in $\mathcal{D}_{\tau}\cup\{\tau\}$ share the same weights.

A target particle will not be tagged as dead at all, i.e. $\tau \not\in \mathcal{D}_k$ for any dead particle set $\mathcal{D}_k$, since a target particle is drawn according to the particles' weights and since dead particles are defined as the ones having too small or zero weight to be chosen. Hence, Equation~\ref{eq:re_correct_1} can be rewritten as
\begin{equation}
    \sum_{i\in\mathcal{D}_\tau} e^{L_i(s_t)} + e^{L_\tau(s_t)} = (\vert\mathcal{D}_\tau\vert+1) e^{L_\tau^\prime(s_t)}, 
\end{equation}

Given that $e^{L_i(s_t)} \approx 0$ for any dead particle $i \in \mathcal{D}_\tau$ and that the number of particles is limited, it implies that
\begin{equation}
    e^{L_\tau} \approx (\vert\mathcal{D}_\tau\vert+1) e^{L_\tau^\prime(s_t)}. 
\end{equation}
Taking the logarithm of both sides of the equation leads to that for all particles in $\mathcal{D}_{\tau}\cup\{\tau\}$, their new logits after resampling should satisfy
\begin{equation}
    L_\tau^\prime(s_t) \approx L_\tau(s_t) - \log(\vert\mathcal{D}_\tau\vert+1).
\end{equation}

Assuming the input of the full-connected layer who generates $L_i$ is $\mathbf{x}(s_t)$, i.e. $L_i(s_t) = \boldsymbol\omega_i\mathbf{x}(s_t) + b_i$, we have
\begin{equation}
    \boldsymbol\omega_i^\prime \mathbf{x}(s_t) + b_i^\prime = \boldsymbol\omega_{\tau} \mathbf{x}(s_t) + b_{\tau} - \log\left(\vert\mathcal{D}_{\tau}\vert + 1\right).
\end{equation}

Then, Theorem can be reached.
\end{proof}

If we perform unweighted resampling, it is possible to pick a dead particle as the target particle for some particles. In that case
\begin{equation}
    L_\tau^\prime(s_t) \approx L_\tau(s_t) - \log(\vert D_\tau \vert + (1-\sum_k \delta(\tau, \mathcal{D}_k))),
\end{equation}
where $L_\tau^\prime(s_t)$ is the new logits shared by particles in $\mathcal{D}_\tau$ and $\delta(\tau, \mathcal{D}_k)$ is the Kronecker delta function
\begin{equation}
    \delta(\tau, \mathcal{D}_k) = \left\{\begin{array}{ll}
         1 & \textrm{if } \tau \in \mathcal{D}_k   \\
         0 & \textrm{otherwise}
    \end{array}
    \right.
\end{equation}
that satisfies $\sum_k \delta(\tau, \mathcal{D}_k) \leq 1$. Then, for the particle $\tau$, its new logits can be defined as
\begin{equation}
    L_\tau^{\prime\prime}(s_t) \approx (1-\sum_k \delta(\tau, \mathcal{D}_k)) L_\tau^\prime(s_t) + \sum_k \delta(\tau, \mathcal{D}_k) L_\tau. 
\end{equation}
Consequently, the target particle $\tau$ may or may not share the same logits with those in $\mathcal{D}_\tau$, depending on if it is tagged as dead or not.

\section{Variance of Policy Gradient in PFPN Configuration}\label{app:grad_var}
Since each action dimension is independent to others, without loss of generality, we here consider the action $a_t$ with only one dimension along which $n$ particles are distributed and the particle $i$ to represent a Gaussian distribution $\mathcal{N}(\mu_i, \sigma_i^2)$. In order to make it easy for analysis, we set up the following assumptions: the reward estimation is constant, i.e. $A_t \equiv A$; logits to support the weights of particles are initialized equally, i.e. $w_i(s_t \vert \theta) \equiv \frac{1}{n}$ for all particles $i$ and $\nabla_\theta w_1(\mathbf{s}_t \vert \theta) = \cdots = \nabla_\theta w_n(\mathbf{s}_t \vert \theta)$; particles are initialized to equally cover the whole action space, i.e. $\mu_i=\frac{i-n}{n}$, $\sigma_i^2 \approx \frac{1}{n^2}$ where $i=1,\cdots,n$.

From Equation~\ref{eq:policy_update}, the variance of the policy gradient under such assumptions is
\begin{equation}\label{eq:variance}\begin{array}{rl}
\mathbb{V}[\nabla_\theta J(\theta) \vert a_t] & = \int \frac{A_t \sum_i p_i(a_t \vert \mu_t, \sigma_t) \nabla_\theta w_i(\mathbf{s}_t\vert\theta)}{\sum_i w_i(s_t \vert \theta) p_i(a_t \vert \mu_t, \sigma_t)} a_t^2 \mathrm{d}a_t  \\ [5pt]
     & \propto \sum_i \nabla_\theta w_i(\mathbf{s}_t\vert\theta) \int a_t^2 p_i(a_t \vert \mu_t, \sigma_t) \mathrm{d}a_t \\ [5pt]
     & \stackrel{\propto}{\sim}  \sum_i (\mu_i^2 + \sigma_i^2) \nabla_\theta w_i(\mathbf{s}_t\vert\theta) \\ [5pt]
     & \propto \sum_i \frac{(i-n)^2+1}{n^2} \\ [5pt]
     & = \frac{n}{3}+\frac{7}{6n}-\frac{1}{2} \\ [5pt]
     & \sim 1 - \frac{3}{2n} + O(\frac{1}{n^2}).
\end{array}\end{equation}
Given $\mathbb{V}[\nabla_\theta J(\theta) \vert a_t] = 0$ when $n=1$, from Equation~\ref{eq:variance}, for any $n > 0$, the variance of policy gradient $\mathbb{V}[\nabla J(\theta) \vert a_t]$ will increase with $n$. Though the assumptions usually are hard to meet perfectly in practice, this still gives us an insight that employing a large number of particles may result in more challenge to optimization.

This conclusion is consistent with that in the case of uniform discretization~\cite{tang2019discretizing} where the variance of policy gradient is shown to satisfy
\begin{equation}
    \mathbb{V}[\nabla_\theta J(\theta) \vert a_t]_{\textsc{discrete}} \sim 1 - \frac{1}{n}.
\end{equation}

That is to say, in either PFPN or uniform discretization scheme, we cannot simply improve the control performance of the police by employing more atomic actions, i.e. by increasing the number of particles or using more bins in the uniform discretization scheme, since the gradient variance increases as the discretization resolution increases. However, PFPN has a slower increase rate,
which implies that it might support more atomic actions before performance drops due to the difficulty in optimization. Additionally, compared to the fixed, uniform discretization scheme, atomic actions represented by particles in PFPN are movable and their distribution can be optimized. This means that PFPN has the potential to provide better discretization scheme using fewer atomic actions to meet the fine control demand and thus be more friendly to optimization using policy gradient.

\section{PFPN with Off-Policy Policy Gradient Algorithms}\label{app:reparameter}
To enable PFPN applicable in state-action value based off-policy algorithms, we propose a reparamterization trick in this section such that a sampled action $a_\theta^{\mathcal{P}}(\mathbf{s}_t)$ can be differentiable to the policy network parameter $\theta$.

\subsection{Reparameterization Trick}

Let $\mathbf{x}(\mathbf{s}_t\vert\theta)\sim\textsc{Concrete}(\{w_i(\mathbf{s}_t\vert\theta); i=1,2,\cdots\}, \lambda)$
is a sampling result of a relaxed version of the one-hot categorical distribution supported by the probability of $\{w_i(\mathbf{s}_t\vert\theta); i=1,2,\cdots\}$,
where $\mathbf{x}(\mathbf{s}_t\vert\theta) = \{x_i(\mathbf{s}_t\vert\theta);i=1,2,\cdots\}$ is reparametrizable and $\lambda$ is picked to be $1$ in our implementation.
We apply the Gumbel-softmax trick~\cite{jang2017categorical} to get a sampled action value as
\begin{equation}
    a^\prime(\mathbf{s}_t) = \textsc{stop}\left(\sum_i a_i \delta(i, \arg\max \mathbf{x}(\mathbf{s}_t\vert\theta))\right),
\end{equation}
where
$a_i$ is the sample drawn from the distribution represented by the particle $i$ with parameter $\phi_i$,
$\textsc{stop}(\cdot)$ is a ``gradient stop'' operation,
and $\delta(\cdot, \cdot)$ denotes the Kronecker delta function. 
Then, the reparameterized sampling result can be written as follows:
\begin{equation}
a_\theta^{\mathcal{P}}(\mathbf{s}_t) = \sum_i (a_i - a^\prime(\mathbf{s}_t))m_i + a^\prime(\mathbf{s}_t)\delta(i, \arg\max \mathbf{x})
\equiv a^\prime(\mathbf{s}_t),
\end{equation}
where $m_i := x_i(\mathbf{s}_t\vert\theta) + \textsc{stop}(\delta(i, \arg\max \mathbf{x}(\mathbf{s}_t\vert\theta)) - x_i(\mathbf{s}_t\vert\theta)) \equiv \delta(i, \arg\max \mathbf{x}(\mathbf{s}_t\vert\theta))$ composing a one-hot vector that approximates the samples drawn from the corresponding categorical distribution.
Since $x_i(\mathbf{s}_t\vert\theta)$ drawn from the concrete distribution is differentiable to the parameter $\theta$, the gradient of the reparameterized action sample can be obtained by
\begin{equation}\begin{array}{l}
    \nabla_{\theta} a_\theta^{\mathcal{P}}(\mathbf{s}_t) = \sum_i(a_i - a^\prime(\mathbf{s}_t))\nabla_{\theta}x_i(\mathbf{s}_t\vert\theta); \\ [5pt]
    \nabla_{\phi_i} a_\theta^{\mathcal{P}} = \delta(i, \arg\max \mathbf{x}(\mathbf{s}_t\vert\theta))\nabla_{\phi_i}a_i.
\end{array}
\end{equation}
Through these equations, both the policy network parameter $\theta$ and the particle parameters $\phi_i$ can be updated by backpropagation through the sampled action $a^\prime(\mathbf{s}_t)$.

\subsection{Policy Representation with Action Bounds}
In off-policy algorithms, like DDPG and SAC, an invertible squashing function, typically the hyperbolic tangent function, will be applied to enforce action bounds on samples drawn from Gaussian distributions, e.g. in SAC,
the action for the $k$-th dimension at the time step $t$ is obtained by
\begin{equation}
    a_{t,k}(\varepsilon, \mathbf{s}_t) = \tanh u_{t,k}
\end{equation}
where $u_{t,k} \sim \mathcal{N}(\mu_\theta(\mathbf{s}_t), \sigma_\theta^2(\mathbf{s}_t))$, $\mu_\theta(\mathbf{s}_t)$ and $\sigma_\theta^2(\mathbf{s}_t)$ are parameters generated by the policy network with parameter $\theta$, and $u_{t,k}$ can be written $u_{t,k} = \mu_\theta(\mathbf{s}_t) + \xi_{t,k} \sigma_\theta^2(\mathbf{s}_t)$ given a noise variable $\xi_{t,k}\sim\mathcal{N}(0, 1)$ such that $a_{t,k}$ is reparameterizable.

Let $\mathbf{a}_t = \{\tanh u_{t,k}\}$ where $u_{t,k}$, drawn from the distribution represented by a particle with parameter $\phi_{t,k}$, is a random variable sampled to support the action on the $k$-th dimension. Then, the probability density function of PFPN represented by Equation~\ref{eq:particle_policy} can be rewritten as
\begin{equation}
    \pi_{\theta}^{\mathcal{P}}(\mathbf{a}_t \vert \mathbf{s}_t) = \prod_k \sum_i w_{i,k}(\mathbf{s}_t \vert \theta) p_{i,k}(u_{t,k} \vert \phi_{i,k})/(1-\tanh^2 u_{t,k}),
\end{equation}
and the log-probability function becomes
\begin{equation}\begin{array}{rl}
     \log \pi_{\theta}^{\mathcal{P}}(\mathbf{a}_t \vert \mathbf{s}_t)&  = \sum_k \log \left[\sum_i w_{i,k}(\mathbf{s}_t \vert \theta) p_{i,k}(u_{t,k} \vert \phi_{i,k})\right. \\ [5pt]
     &  \left. - 2\left(\log2 - u_{t,k} - \mathrm{softplus}(-2u_{t,k})\right) \right].
\end{array}
\end{equation}

In our SAC implementation, we use Gaussian noises to generate action samples, i.e. $u_{i,k} \sim \mathcal{N}(\mu_{i,k}, \xi_{i,k}^2)$ where $\mu_{i,k}$ and $\xi_{i,k}$ are the parameters for the $i$-th particle at the $k$-th action dimension.

\end{document}